\documentclass[11pt,a4paper]{article}
\usepackage[hyperref]{emnlp2020}
\usepackage{times}
\usepackage{latexsym}

\usepackage{microtype}
\usepackage{graphics} 
\usepackage{graphicx} 
\usepackage{amsmath} 
\usepackage{amsfonts}
\usepackage{subcaption}
\usepackage[nopar]{lipsum}
\usepackage{enumitem} 
\setlist[itemize]{noitemsep, topsep=0pt}
\setlist[enumerate]{noitemsep, topsep=0pt}
\usepackage{comment} 
\usepackage{multirow}
\usepackage{rotating} 
\usepackage{xcolor}
\usepackage{colortbl}
\usepackage{textgreek}
\usepackage{lettrine}

\usepackage{url}
\aclfinalcopy 
\def\aclpaperid{2718} 

\title{Bridging Linguistic Typology and Multilingual Machine Translation \\ with Multi-View Language Representations}

\author{Arturo Oncevay \quad Barry Haddow \quad Alexandra Birch \\
  School of Informatics, University of Edinburgh, Scotland \\
  \texttt{\{a.oncevay,a.birch\}@ed.ac.uk,bhaddow@inf.ed.ac.uk} \\
  }

\date{}

\begin{document}
\maketitle
\begin{abstract}
  Sparse language vectors from linguistic typology databases and learned embeddings from tasks like multilingual machine translation have been investigated in isolation, without analysing how they could benefit from each other's language characterisation. We propose to fuse both views using singular vector canonical correlation analysis and study what kind of information is induced from each source. By inferring typological features and language phylogenies, we observe that our representations embed typology and strengthen correlations with language relationships. We then take advantage of our multi-view language vector space for multilingual machine translation, where we achieve competitive overall translation accuracy in tasks that require information about language similarities, such as language clustering and ranking candidates for multilingual transfer. With our method, which is also released as a tool, we can easily project and assess new languages without expensive retraining of massive multilingual or ranking models, which are major disadvantages of related approaches.
\end{abstract}

\setlength{\abovedisplayskip}{3pt}
\setlength{\belowdisplayskip}{3pt}

\section{Introduction}
Recent surveys consider linguistic typology as a potential source of knowledge to support multilingual natural language processing (NLP) tasks \cite{ohoran-etal-2016-survey,ponti-etal-2019-modeling}. 
Linguistic typology studies language variation in terms of their functional processes 
\cite{comrie1989language}. Several typological knowledge bases (KB) have been crafted, from where we can extract categorical language features \cite{littell-etal-2017-uriel}. Nevertheless, their sparsity and reduced coverage 
present a challenge for an end-to-end integration into NLP algorithms. 
For example, the World Atlas of Language Structure~\cite[WALS;][]{wals} encodes 143 features for 2,679 languages, but their mean coverage per language is barely around 14\%.  

Dense and data-driven language representations have emerged in response. They are computed from multilingual settings of language modelling \cite{ostling-tiedemann-2017-continuous} and neural machine translation (NMT) \cite{malaviya-etal-2017-learning}. 
However, the language diversity in the corpus-based representations is limited. The language coverage could be broadened with other knowledge, such as that encoded in WALS, to distinguish even more language properties. 
Therefore, to obtain the best of both views (KB and task-learned) with minimal information loss, 
we project a shared space of discrete and continuous features using a variant of canonical correlation analysis \cite{svcca-NIPS2017_7188}. 

For our study, we fuse language-level embeddings from multilingual machine translation with syntactic features of WALS. We inspect how much typological knowledge is present by predicting features for new languages. Then, we infer language phylogenies and inspect whether specific relationships are induced from the task-learned vectors.   

Furthermore, to demonstrate that our approach has practical benefits in NLP, we apply our language vectors in multilingual NMT with language clustering \cite{tan-etal-2019-multilingual} and adapt the ranking of related languages for multilingual transfer \cite{lin-etal-2019-choosing}. 
As a side outcome, we identify that there is an ideal setting to encode language relationships in language embeddings from NMT. Finally, we are releasing a simple tool to allow everyone to fuse their own representations for clustering, ranking and more.

\section{Multi-view language representations} 
\label{sec:multi-langrep}
Our primary goal is to fuse parallel representations of the same language in one shared space, and \textbf{canonical correlation analysis} (CCA) 
allows us to find a projection of two views for a given set of data. 
With CCA, we look for linear combinations that maximise the correlation of the two sources in each coordinate iteratively \cite{hardoon2004canonical}. 
After training, we can apply the transformation learned on a new sample from any view to obtain a CCA-based language representation.\footnote{With language representations, we refer to an annotated or unsupervised characterisation of a language itself (e.g. Spanish or English), and not to word or sentence-level representations, as it is used in the recent NLP literature.}

CCA considers all dimensions of the two views as equally important. However, our sources are potentially redundant: KB features are mostly one-hot-encoded, whereas task-learned ones inherit the high dimensionality of the embedding layer. Moreover, few samples and sparsity could make the convergence harder.
For the redundancy issue, \textbf{singular value decomposition} (SVD) is an appealing alternative. 
With SVD, we factorise the source data matrix to compute the principal components and singular values. Furthermore, to deal with sparsity, we adopt a truncated SVD approximation, which is also known as latent semantic analysis in the context of linear dimensionality reduction for term-count matrices \cite{dumais2004latent}. 

The two-step transformation of SVD followed by CCA is called \textbf{singular vector canonical correlation analysis}~\cite[SVCCA;][]{svcca-NIPS2017_7188} in the context of understanding the representation learning throughout neural network layers. That being said, we use SVCCA to get language representations and not to inspect a neural architecture.\footnote{As the SVD step performs a dimensionality reduction while preserving the most explained variance as possible, we can consider two additional parameters: a threshold value in the [0.5,1.0] range with 0.05 incremental steps, for the explained variance ratio of each view. With a value equal to 1, we bypass SVD and compute CCA only. We then tuned all our following experiments (see Appendix \ref{app:parameter-search} for details).}

\section{Methodology and research questions}

To embed linguistic typology knowledge in dense representations for a broad set of languages, we employ SVCCA (\S\ref{sec:multi-langrep}) with the following sources:

\paragraph{KB view. }
We employ the language vectors from the URIEL and $\operatorname{lang2vec}$ database \cite{littell-etal-2017-uriel}. Precisely, we work with the $k$-NN vectors of the Syntax feature class ($U_S$; 103 feats.), 
that are composed of binary features encoded from WALS~\cite{wals}.

\paragraph{(NMT) Learned view. }
\label{par:nmt-learned}

Firstly, we exploit the NMT-learned embeddings from the Bible ($L_B$; 512 dim.) \cite{malaviya-etal-2017-learning}. Up to 731 entries are available in $\operatorname{lang2vec}$ that intersects with $U_S$. They were trained in a many-to-English NMT model with a pseudo-token identifying the source language at the beginning of every input sentence. 

Secondly, we take the many-to-English language embeddings learned for the language clustering task on multilingual NMT ($L_W$; 256 dim.) \cite{tan-etal-2019-multilingual}, where they use 23 languages of the WIT$^3$ corpus~\cite{cettoloEtAl:EAMT2012}. 

One main difference for the latter is the use of factors in the architecture, meaning that the embedding of every input token was concatenated with the embedded pseudo-token that identifies the source language. The second difference is the neural architecture used to extract the embeddings: the former use a recurrent neural network, whereas the latter a small transformer model \cite{NIPS2017_7181}. 

Finally, we train a new set of embeddings ($L_T$) that we extracted from the 53 languages of the TED corpus (many-to-English) processed by \citet{qi-etal-2018-pre}, using the approach of \citet{tan-etal-2019-multilingual}.\footnote{We prefer to use factored embeddings over initial pseudo-tokens as we identified that there is a difference for encoding information about language similarity (see \S\ref{sec:factors2}).}

\paragraph{What knowledge do we represent?} Each source embeds specialised knowledge to assess language relatedness. The KB vectors can measure typological similarity, whereas task-learned embeddings correlates with other kinds of language relationships (e.g. genetic) \cite{Bjerva2019-bl}. To analyse whether each kind of knowledge is induced with SVCCA, we assess the tasks of typological feature prediction (\S\ref{sec:typ-prediction}) and reconstruction of a language phylogeny (\S\ref{sec:tree-inference}). 

\paragraph{What is the benefit for multilingual NMT (and NLP)?} Language-level representations can evaluate the distance between languages in a vector space. We then can assess their applicability on multilingual NMT tasks that require guidance from language relationships. Therefore, language clustering and ranking related partner languages for (multilingual) transfer are our study cases (\S\ref{sec:broadening-clustering}).

\section{Prediction of typological features}
\label{sec:typ-prediction}

An example of a typological feature is a word order specification, like whether the adjective is predominately placed before or after the noun (features \#24 and \#25 of $U_S$). Our task consists in predicting syntactic features ($U_S$) leaving one-language and one-language-family out to control phylogenetic relationships \cite{bjerva-etal-2019-probabilistic}. Previous work has shown that task-learned embeddings are potential candidates to predict features of a linguistic typology KB \cite{malaviya-etal-2017-learning}, and our goal is to evaluate whether SVCCA can enhance the NMT-learned language embeddings with typological knowledge from their KB parallel view. 

\paragraph{Experimental setup.} We use a Logistic Regression classifier per $U_S$ feature, which is trained with the NMT-learned or SVCCA representations in both one-language-out and one-language-family-out settings. For prediction, we use the original embedding or its SVCCA projection as inputs.

\paragraph{Results.} In Table \ref{tab:typ-prediction}, we observe that SVCCA outperformed their NMT-learned counterparts for $L_W$ and $L_T$, where the performance is significantly better for the one-language-out setting. In the case of $L_B$ (with 731 entries), we notice that the overall performance drops, and the SVCCA transformation cannot improve it. We argue that a potential reason for the accuracy dropping is the method used to extract the NMT-learned embeddings (initial pseudo-token instead of factors: \S\ref{sec:factors2}), which could diminishes the information embedded about each language, and consequently, impacts the SVCCA projection.\footnote{In other words, for SVCCA, it is difficult to deal with the noise provided in the learned embeddings. In Figures \ref{fig:ddg-bible} and \ref{fig:ddg-ted53-pseudotoken} of the Appendix, we observe noisy agglomerations in the dendrograms (obtained by clustering different language representations), which is preserved after the fusing with the KB vectors through SVCCA as we can see in Fig. \ref{fig:ddg-svcca53-pseudotoken})} In conclusion, we notice that specific typological knowledge is usually hard to learn in an unsupervised way, and fusing them with KB vectors using SVCCA is feasible for inducing information of linguistics typology in some scenarios.

\begin{table}[t!]
\centering
\setlength\tabcolsep{4pt}
\resizebox{\linewidth}{!}{%
\begin{tabular}{lcc|ccc}
                     & \multicolumn{2}{c}{one-language-out} & \multicolumn{3}{c}{one-family-out} \\ 
                     & Single & SVCCA  & \#F. & Single & SVCCA  \\ \hline
$L_B$ (Bible)   & \textbf{72.77} & 71.68 & 134 & \textbf{72.15} & 70.62 \\ 
$L_W$ (WIT-23)  & 81.27 & \textbf{84.83} &  12 & 79.49 & \textbf{79.68} \\ 
$L_T$ (TED-53)  & 77.96 & \textbf{85.37} &  18 & 76.36 & \textbf{81.06} \\ \hline
\end{tabular}
}
\caption{Avg. accuracy ($\uparrow$) of typological feature prediction per NMT-learned and SVCCA($U_S$,$L_*$) setting.} 

\label{tab:typ-prediction}
\end{table}

\section{Language phylogeny analysis}
\label{sec:tree-inference}

According to \citet{Bjerva2019-bl}, there is a positive correlation between the language distances in a phylogenetic tree and a pairwise distance-matrix of task-learned representations. Our goal therefore is to investigate whether fusing linguistic typology with SVCCA can preserve or enhance the embedded relationship information. For that reason, we examine how well a language phylogeny can be reconstructed from language representations (\S\ref{subsec:infer-tree}), and also study the correlation (Appendix \ref{app:lexical-sim}).

\subsection{Inference of a phylogenetic tree}
\label{subsec:infer-tree}

\begin{figure}[t!]
\begin{center}

\begin{subfigure}[t]{0.2\textwidth}
\caption{Gold Standard} 
\label{fig:tree-gs}
\includegraphics[width=\linewidth,trim={0 0cm 0 0.375cm},clip]{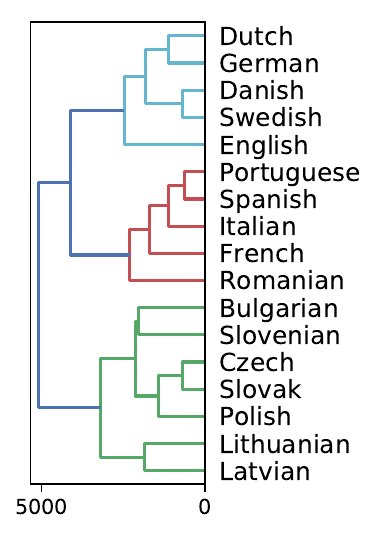}
\end{subfigure}
\begin{subfigure}[t]{0.2\textwidth}
\caption{SVCCA($U_S$,$L_T$)}
\label{fig:tree-svcca}
\includegraphics[width=\linewidth,trim={0 0cm 0 0.375cm},clip]{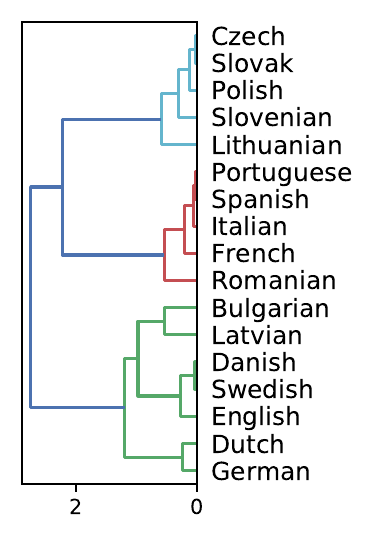} 
\end{subfigure}
\begin{subfigure}[t]{0.2\textwidth}
\caption{$U_S$: Syntax}
\label{fig:tree-syntax}
\includegraphics[width=\linewidth,trim={0 0.3cm 0 0.375cm},clip]{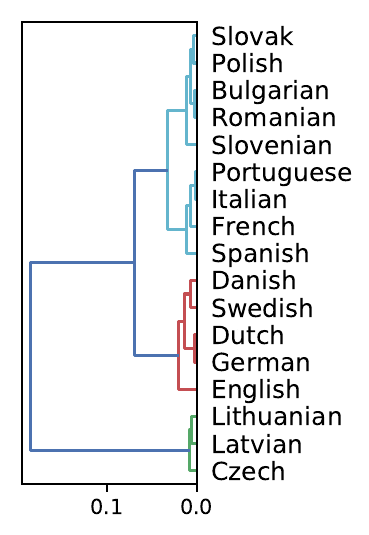} 
\end{subfigure}
\begin{subfigure}[t]{0.2\textwidth}
\caption{$L_T$: TED-53}
\label{fig:tree-ted53}
\includegraphics[width=\linewidth,trim={0 0.3cm 0 0.375cm},clip]{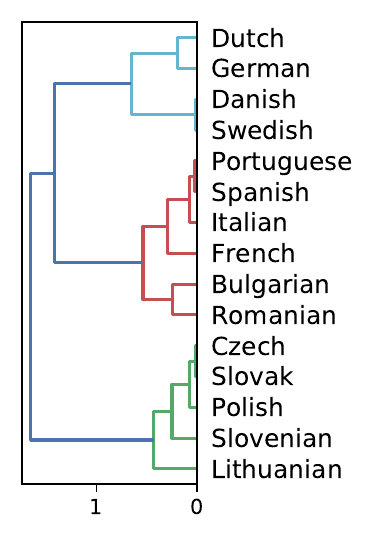}
\end{subfigure}

\caption{Gold Standard phylogeny (a) and reconstructed trees (b-d). $L_T$ is smaller.}
\label{fig:GS-trees}
\end{center}
\end{figure}

\paragraph{Experimental design.} Based on previous work \cite{rabinovich-etal-2017-found}, we take a tree of 17 Indo-European languages \cite{Serva2008-fb} as a Gold Standard (GS), which is shown in Figure \ref{fig:GS-trees}a.\footnote{We do not generalise the analysis for more languages, as the inferred tree of \citet{Serva2008-fb} is only an approximation by lexicostatistic methods (see Appendix \ref{app:lexical-sim}).} We also use agglomerative clustering with variance minimisation \cite{ward1963hierarchical} as linkage, but we employ cosine similarity as \citet{Bjerva2019-bl}.

We also consider a concatenation ($\oplus$) of the KB and NMT-learned views as a baseline. 

It is essential to highlight that none of the NMT-learned and $\oplus$ vectors have all the 17 language entries of the GS. Therefore, we can already see one of the significant advantages of the SVCCA vectors, as we are able to represent ``unknown'' languages using one of the views. The NMT-learned views lack English, since they were extracted from the source side of a many-to-English system, but we were able to project the KB English vectors into the shared space.\footnote{This is illustrative only, as we could obtain an English vector from many-to-many multilingual NMT models or language models. However, the artificial case generalises as a benefit for projecting new languages with SVCCA. For instance, $\operatorname{lang2vec}$ contains 2,989 and 287 unique entries in the KB and NMT-learned views, respectively.} In addition, we project other four languages (Swedish, Danish, Latvian, Lithuanian) to complete the $L_W$ embeddings of \citet{tan-etal-2019-multilingual} and Latvian to complete our own $L_T$ set.

\paragraph{Evaluation metric.} We differ from previous studies and use a tree edit distance metric, which is defined as the minimum cost of transforming one tree into another by inserting, deleting or modifying (the label of) a node. Specifically, we used the All Path Tree Edit Distance algorithm \cite[APTED;][]{pawlik-augsten-2015-efficient,pawlik-augsten-2016-tree}, a novel one for the task. We chose an edit-distance method as it is more transparent for assessing what is the degree of impact for a single change of linkage in the GS.

As we need to compare inferred pruned trees with different number of nodes, we propose a normalised version given by: $\operatorname{nAPTED} = \operatorname{APTED} / (|\operatorname{GS}| + |\tau|)$, where $\tau$ is the inferred tree, and $|.|$ indicates the number of nodes. The denominator then is the maximum cost possible of deleting all nodes of $\tau$ and inserting each GS node.

\begin{table}[t!]
\centering
\setlength\tabcolsep{4pt}
\resizebox{\linewidth}{!}{%
\begin{tabular}{l|c|cc}
                     & KB-view \\    \hline
$U_S$ (Syntax)  & 30 -- 0.45 \\    \hline
                     & NMT-learned & $U_S\oplus L_*$  & SVCCA($U_S$,$L_*$)  \\ \hline
$L_B$ (Bible)   & 35 -- 0.54 & 27 -- 0.42 & 23 -- 0.34  \\
$L_W$ (WIT-23)  & 35 -- 0.62 & 23 -- 0.41 & 27 -- 0.48  \\
$L_T$ (TED-53)  & 15 -- 0.26 & 18 -- 0.29 & \textbf{10} -- \textbf{0.15} \\ \hline
\end{tabular}
}
\caption{APTED and nAPTED scores ($\downarrow$) between the GS and inferred trees from all scenarios. APTED ranges from 0 (no difference) and the size of the tree at most. NMT-learned and concatenation ($\oplus$) can only reconstruct pruned trees of 16 ($L_B$), 12 ($L_W$) and 15 ($L_T$) languages. }
\label{tab:tree-TED-multi-views}
\end{table}

\paragraph{Results.} Table \ref{tab:tree-TED-multi-views} shows the results for all settings,  
where the single-view scores are meagre in most of the cases. For instance, the $U_S$ inferred tree (Fig.\ref{fig:GS-trees}c) requires 30 edits to match the GS. The exception is $L_T$ (Fig.\ref{fig:GS-trees}d), which requires half the edits, although it is incomplete.

 We observe that the best absolute and normalised scores are obtained by fusing $U_S$ and $L_T$ with SVCCA (Fig.\ref{fig:GS-trees}b). English is projected in the Germanic branch, although Latvian is separated from the Balto-Slavic group. The latter case is similar for Bulgarian, which is misplaced in the original $L_T$ tree as well. Nevertheless, we only require ten editions to equate the GS (where 66 is the maximum cost possible), confirming that our approach is a robust alternative for completing language entries and inferring a language phylogeny.\footnote{In further analysis, we confirmed that the inferred tree with only 12 languages of SVCCA (without projection of extra entries) is comparable or better against the rest of the baselines.}
 
 In conclusion, we observe that using typological knowledge with SVCCA enhances the language relationship encoded in the NMT-learned embeddings. In Appendix \ref{app:lexical-sim}, we further discuss what kind of relationship we are representing in the NMT-learned embeddings and SVCCA, and study their correlation.

\section{Application in multilingual NMT}
\label{sec:broadening-clustering}

With multilingual NMT, we can translate several language-pairs using a single model. Low-resource languages  usually benefit through multilingual transfer, which resembles a simultaneous training of the parent(s) and child models. Therefore, 
we want to take advantage of a language-level vector space for relating similar languages and enhancing multilingual transfer within multilingual NMT. For that reason, we first address the language clustering task proposed by \citet{tan-etal-2019-multilingual}, and afterwards, the language ranking model of \citet{lin-etal-2019-choosing}. 

\paragraph{Language clustering.} The main idea is to obtain smaller multilingual NMT models as an intermediate point between maintaining many pairwise systems and a single massive multilingual model. With limited resources, it is challenging to support the first scenario, whereas the advantages for the massive setting are also very appealing (e.g. simplified training process, translation improvement for low-resource languages or zero-shot translation \cite{johnson-etal-2017-googles}). Therefore, to address the task, \citet{tan-etal-2019-multilingual} trained a factored multilingual NMT model of 23 languages from \citet{cettoloEtAl:EAMT2012}, where the language embedding is concatenated in every input token. 
Then, they performed hierarchical clustering with the representations, and selected a number of clusters guided by the Elbow method. Finally, they compared the systems against individual, massive and language family-based cluster models. 

In a practical multilingual NMT system, it is not only necessary to choose the right clustering, the ability to easily add new languages is also important.
With this in mind, we apply our multi-view representations to compute a set of clusters, and we also address the question: do we need to train the massive model again if we want to add one or more new languages to our setting?  

\paragraph{Language ranking.} The original goal of \textsc{LangRank} is to choose a parent language to perform transfer learning in different tasks, NMT included. To achieve this, \citet{lin-etal-2019-choosing} trained a model based on the performance of several hundred pairwise MT systems using the dataset of \citet{qi-etal-2018-pre}. For the input features, they considered linguistically-informed vectors from $\operatorname{lang2vec}$ \cite{littell-etal-2017-uriel} and corpus-based statistics, such as word/sub-word overlapping and the ratio of the token-types or the data size between the target child and potential candidates, where the latter features were some of the most relevant. 

Considering the transfer capabilities within multilingual NMT and the possibility to obtain a ranked list of candidates from \textsc{LangRank}, we propose an adapted task of choosing $k$-related languages for multilingual transfer. We then use our multi-view representations to rank related languages from the vector space, as they embed information about typological and lexical relationships. This is similar to the features that \citet{lin-etal-2019-choosing} consider, but without training a ranking model fed with scores from pairwise MT systems.

\subsection{Experimental setup}
\label{sec:clust-setup}

We focus on the many-to-one (English) multilingual NMT setting to simplify the evaluation in both tasks. However, similar experiments could be performed in a one-to-many direction. 

\paragraph{Dataset. }
We use the dataset processed and tokenised by \citet{qi-etal-2018-pre} of 53 languages (TED-53), from where we learned our $L_T$ embeddings. We opted for TED-53 to better evaluate the extensibility of clusters and because it is also used to train the \textsc{LangRank} model. The list of languages, set sizes and other details are included in Appendix \ref{app:languages}. Before preprocessing the text, we drop any sentences from the training sets which overlap with any of the test sets. Since we are building many-to-English multilingual systems, this is important, as any such overlap will bias the results. 

\paragraph{Model and training.}
Similar to \citet{tan-etal-2019-multilingual}, we train small transformer models \cite{NIPS2017_7181}. 
We jointly learn 90k shared sub-words with the byte pair encoding \cite{sennrich-etal-2016-neural} algorithm built in SentencePiece \cite{kudo-richardson-2018-sentencepiece}. We also oversample all the training data of the less-resourced languages in each cluster, and shuffle them proportionally in all batches. 
 
 We use Nematus \cite{sennrich-EtAl:2017:EACLDemo} only to extract the factored language embeddings from the TED-53 corpus ($L_T$). Besides, given the large number of experiments, we also choose the efficient Marian NMT \cite{junczys-dowmunt-etal-2018-marian} toolkit for training the rest of systems. With Marian NMT, we only use the basic pseudo-token setting for identifying the source language, as we did not need to retrieve new language embeddings after training. Besides, we allow the Marian NMT framework to automatically determine the mini-batch size given the sentence-length and available memory (mini-batch-fit parameter). 

We train our models with up to four NVIDIA P100 GPUs using Adam optimiser \cite{kingma2014adam} with default parameters ($\beta_{1}=0.9, \beta_{2}=0.98, \varepsilon=10^{-9}$) and early stopping at 5 validation steps for the cross-entropy metric. Finally, the sacreBLEU version string \cite{post-2018-call} is as follows: BLEU+case.mixed+numrefs.1+smooth.exp +tok.13a+version.1.3.7.

\begin{figure*}[t!]
\begin{center}
\centering

\begin{subfigure}[t]{\linewidth}
\caption{SVCCA-53($U_S,L_T$): SVCCA representations of Syntax and TED-53} 
\label{fig:ddg-svcca53}
\includegraphics[width=\linewidth,clip]{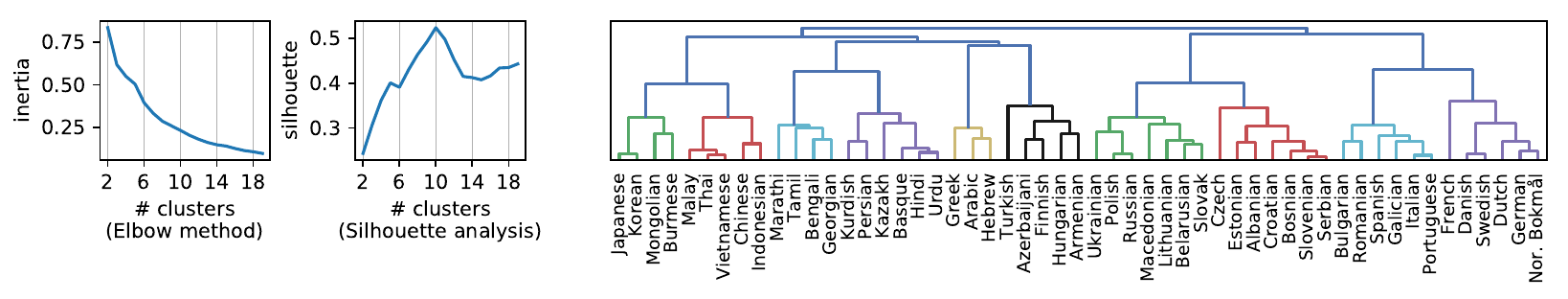}
\end{subfigure}

\begin{subfigure}[t]{\linewidth}
\caption{SVCCA-23($U_S,L_W$): SVCCA representations of Syntax and WIT-23} 
\label{fig:ddg-svcca23}
\includegraphics[width=\linewidth,clip]{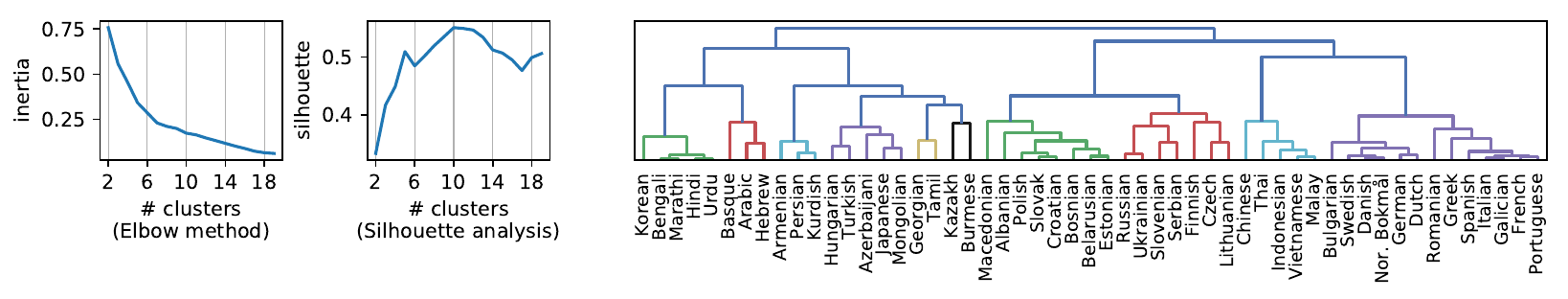}
\end{subfigure}

\begin{subfigure}[t]{\linewidth}
\caption{$U_S$: Syntax} 
\label{fig:ddg-syntax}
\includegraphics[width=\linewidth,clip]{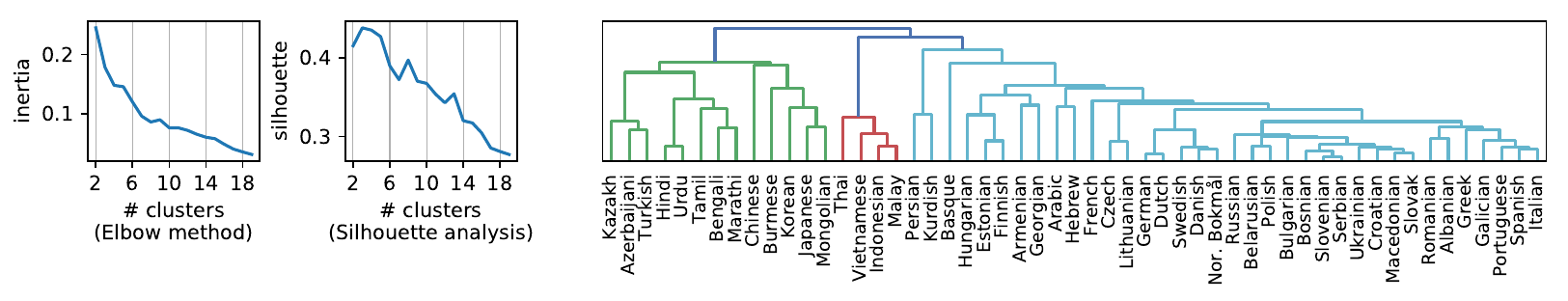}
\end{subfigure}

\begin{subfigure}[t]{\linewidth}
\caption{$L_T$: NMT-learned from TED-53 (using factors)} 
\label{fig:ddg-ted53}
\includegraphics[width=\linewidth,clip]{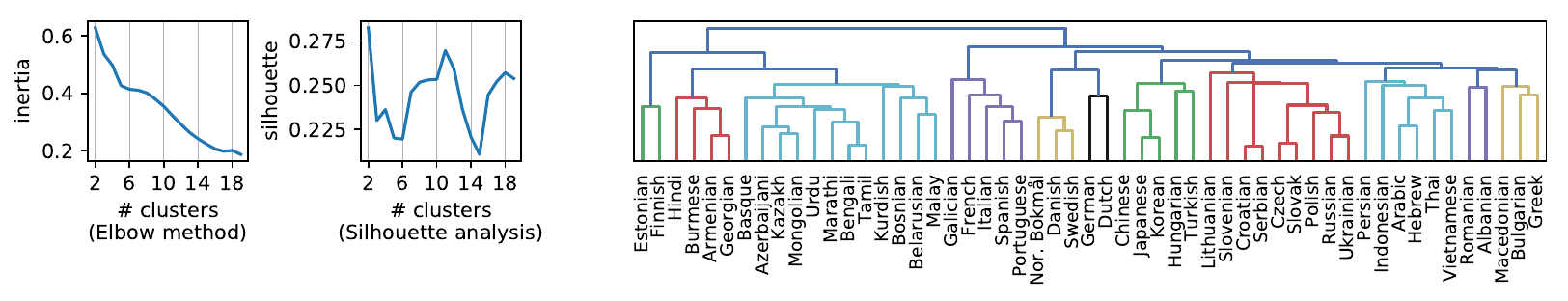}
\end{subfigure}

\caption{(a) Clustering of TED-53 using the SVCCA-53 representations. At the left, we include the Elbow and Silhouette criteria to define the number of clusters. For the former, it is not clear what is the value to choose, whereas for the later we automatically select the highest peak at ten clusters. (b-d) Elbow method, silhouette analysis and dendrograms for SVCCA-23($U_S$,$L_W$) with 30 additionally projected languages, $U_S$ and $L_T$.} 
\label{fig:dendrograms}
\end{center}
\end{figure*}

\paragraph{Clustering settings. } We first list the baselines and our approaches, with the number of clusters/models between brackets:
\begin{enumerate}
    \item Individual [53]: Pairwise model per language.
    \item Massive [1]: A single model for all languages.
    \item Language families [20]: Based on historical linguistics. We divide the 33 Indo-European languages into 7 branches. Moreover, 11 groups only have one language.
    \item KB [3]: $U_S$ (Syntax) tends to agglomerate large clusters (with 4-13-33 languages), behaving similar to a massive model (Fig. \ref{fig:ddg-syntax}).
    \item Learned [11]: We train a set of 53 factored embeddings ($L_T$) similar to \citet{tan-etal-2019-multilingual} (Fig. \ref{fig:ddg-ted53}).
    \item Concatenation [18]: $U_S \oplus L_T$.
    \item SVCCA-53 [10]: Multi-view representation with SVCCA composing both $U_S$ and $L_T$ vectors (Fig. \ref{fig:ddg-svcca53}).
    \item SVCCA-23 [10]: Similar to the previous setting, but we use the set of 23 language embeddings $L_W$ instead \cite{tan-etal-2019-multilingual}, and project the 30 complementary languages with SVCCA($U_S$,$L_W$) (Fig. \ref{fig:ddg-svcca23}). 
\end{enumerate}
With the last setting, we are interrogating whether SVCCA is a useful method for rapidly increasing the number of languages without retraining massive models given new entries that require their NMT-learned embeddings for clustering. 

Similar to \citet{tan-etal-2019-multilingual}, we use hierarchical agglomeration with average linkage and cosine similarity. However, we choose a different criterion for choosing the optimal number of clusters.

\paragraph{Selection of number of clusters. }
The Elbow criterion has been suggested for this purpose~\cite{tan-etal-2019-multilingual}; however, as we can see in Figure \ref{fig:dendrograms}, it might be ambiguous. Thus, we propose using a heuristic called Silhouette \cite{rousseeuw1987silhouettes}, which returns a score in the [-1,1] range. A sample cluster with a silhouette close to 1 indicates that it is cohesive and well-separated. With the average silhouette of all samples, we vary the number of clusters, and look for the peak value above two. 

\paragraph{Ranking settings.} We focus on five low-resource languages from TED-53: Bosnian (bos, Indo-European/Balto-Slavic), Galician (glg, Indo-European/Italic), Malay (zlm, Austronesian), Estonian (est, Uralic) and Georgian (kat, Kartvelian). They have between 5k and 13k translated sentences with English, and we chose them as they achieved the most significant improvement from the individual to the massive setting. We then identified the top-3 related languages using \textsc{LangRank}, which give us a multilingual training set of around 500 thousand sentences for each case.
Given that \textsc{LangRank} usually prefers to choose candidates with larger data size \cite{lin-etal-2019-choosing}, for a fair comparison, we use SVCCA and cosine similarity to choose the $k$ closest languages that can agglomerate a similar amount of parallel sentences.

\subsection{Language clustering results}
\label{sec:clust-res}

We first briefly discuss the composition of clusters obtained by SVCCA. Then, we analyse the results grouped by training size bins. We complement the analysis by family groups in Appendix \ref{app:families}. 

\paragraph{Cluster composition: } In Figure \ref{fig:dendrograms}, we observe that SVCCA-53 (Fig. \ref{fig:ddg-svcca53}) has adopted ten clusters with a proportionally distributed number of languages (the smallest one is Greek-Arabic-Hebrew, and the largest one has seven entries). Moreover, the languages are usually grouped by phylogenetic or geographical criteria. These agglomeration trends are adopted from both the KB (Fig. \ref{fig:ddg-syntax}) and NMT-learned (Fig. \ref{fig:ddg-ted53}) sources.

From a more detailed inspection, there are entries that do not correspond to their respective family branches, although the single-view sources might induce the bias. For instance, the $L_T$ phylogenetic tree (Fig. \ref{fig:GS-trees}d) ``misplaced'' Bulgarian within Italic languages. Nevertheless, the unexpected agglomerations rely on the features encoded in the KB or the NMT learning process, and we expect they can uncover surprising clusters to avoid isolating languages without close relatives (e.g. Basque, or even Japanese as the only Japonic member in the set). Another benefit is noticeable in the SVCCA-23 clusters (Fig. \ref{fig:ddg-svcca23}), which have resemblances with the SVCCA-53 agglomeration despite using only 23 languages to compute the shared space.

\begin{figure}[t!]
\centering

\includegraphics[width=\linewidth,trim={0.15cm 1cm 0.15cm 0.25cm},clip]{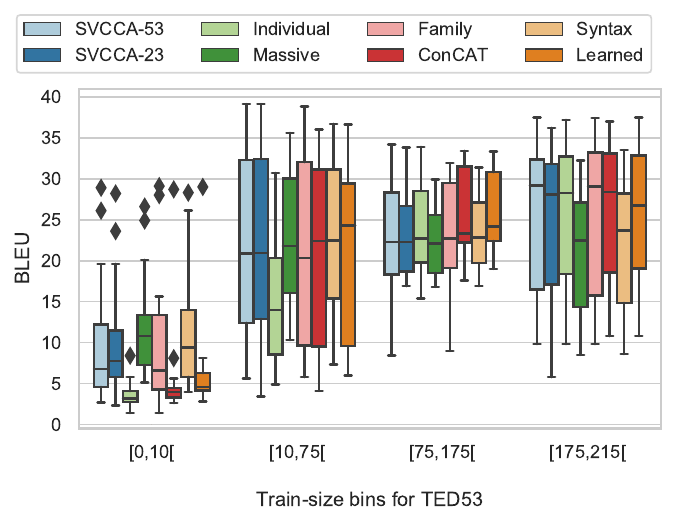}

\caption{Box plots of BLEU scores per training-size bins. Each bin is represented by the range of minimum and maximum training size. Outliers are shown as diamonds.}
\label{fig:binsizes}
\end{figure}

\paragraph{Training size bins: }
We manually define the upper bounds of the bins as [10,75,175,215]  thousands of training sentences, which results in groups composed by [14,14,13,12] languages. Figure \ref{fig:binsizes} shows the box plots of BLEU from where we can analyse each distribution (mean, variance). 

Throughout all the bins, we observe that both SVCCA-53 and SVCCA-23 accomplish a comparable accuracy with the best setting in each group. In other words, their clusters provide stable performance for both low or high-resource languages. 

In the first bin of the smallest corpora, the Massive baseline and the large clusters of $U_S$ barely surpass the SVCCA schemes. Nevertheless, SVCCA contributes a notable advantage if we want to train a multilingual NMT model for a specific low-resource language, and we do not have the resources for training a massive system. We further analyse this scenario in \S\ref{sec:ranking-res}. 

In the rightmost bin, for the highest resource languages, the Massive and $U_S$ performed worse than  SVCCA. Furthermore, we show a competitive accuracy for the Individual and Family approaches. The former's clusters have steady performance across most of the bins as well. Nevertheless, they double the number of clusters that we have in both SVCCA settings, and with more than half of the ``clusters'' having only one language.

Other approaches, like using the NMT-learned embeddings ($L_T$) as \citet{tan-etal-2019-multilingual} or the concatenation baseline, obtain similar translation results in the last three bins. However, we need to obtain the NMT-learned embeddings first in order to fulfil those methods (from a 53-languages massive model). Using SVCCA and a pre-trained smaller set of language embeddings is enough for projecting new representations, as we present with our SVCCA-23 approach.

\subsection{Language ranking results}
\label{sec:ranking-res}

After discussing overall translation accuracy for all the languages, we now focus on five specific low-resource cases and how multilingual transfer enhance their performance. Table \ref{tab:ranking-res} shows the BLEU scores of the translation into English for the smaller multilingual models that group each child language with their candidates ranked by \textsc{LangRank} and our SVCCA-53 representations. 

We also include the results of the individual and massive MT systems. Even when the latter baseline provides a significant improvement over the former, we observe that many of the smaller multilingual models outperform the translation accuracy of the massive system. The result suggests that the amount of data is not the most important confound for supporting multilingual transfer in a low-resource language, which is aligned with the literature \cite{wang-neubig-2019-target}. 

Comparing the two ranking approaches, we observe that SVCCA approximates the performance of \textsc{LangRank} in most of the cases. We note that \textsc{LangRank} prefers related languages with large datasets, as it only requires three candidates to group around half a million training samples, whereas SVCCA suggests to include from three to ten languages to reach a similar amount of parallel sentences. However, increasing the number of languages could impact the multilingual transfer negatively (see the case of Georgian or \emph{kat}), as it is analogous to adding different ``out-of-domain'' samples. To alleviate this, we could bypass candidate languages that do not possess a specific amount of training samples.

We argue that our representations still provides a robust alternative to determine which languages are suitable for multilingual transfer learning. 
The notable advantage is that we do not need to pre-train MT systems from a specific dataset, and we can easily extend the coverage of languages without re-training the ranking model to consider new language entries\footnote{However, we do not answer what multilingual NMT really transfers to the low-resource languages. We left that question for further research, together with optimising the $k$ number of languages or the amount of data per each language.}.

\begin{table}[h!]
\centering
\setlength\tabcolsep{4pt}
\begin{tabular}{c|cc|cc}
L                     & Ind. & Mas. & \textsc{LangRank} & SVCCA-53 \\  \hline
                     
bos  & 4.2 & 26.6 & 28.8 \textsubscript{(434)} &  28.2 \textsubscript{[5]} \\ 
glg  & 8.4 & 24.9 & 27.7 \textsubscript{(443)} & 28.4 \textsubscript{[3]}  \\
zlm  & 4.1 & 20.1 & 21.2 \textsubscript{(463)} & 21.0 \textsubscript{[4]} \\
est  & 5.8 & 13.5 & 13.5 \textsubscript{(533)} & 12.1 \textsubscript{[6]} \\ 
kat  & 5.8 & 14.3 & 13.3 \textsubscript{(499)} & 10.5 \textsubscript{[10]} \\ \hline

\end{tabular}
\caption{BLEU scores (L$\rightarrow$English) for Individual, Massive and ranking approaches. \textsc{LangRank} shows the accumulated training size (in thousands) for the top-3 candidates, whereas with SVCCA we approximate the amount of data and include the number of languages between brackets.} 
\label{tab:ranking-res}
\end{table}

\section{Factors over initial pseudo-tokens}
\label{sec:factors2}

We additionally argue that the configuration used to compute the language embeddings impacts what relationship they can learn. For the analysis, we extract an alternative set of 53 language embeddings ($L_T*$) but using the initial pseudo-token setting instead of factors. Then, we perform a silhouette analysis to identify whether we can build cohesive and well-separated clusters of languages. 

\begin{figure}[t!]
\begin{center}
\centering

\begin{subfigure}[t]{0.475\linewidth}
\caption{$L_T*$} 
\label{fig:sil-ted53pseudo}
\includegraphics[width=\linewidth,trim={5.5cm 0.75cm 20cm 0.675cm},clip]{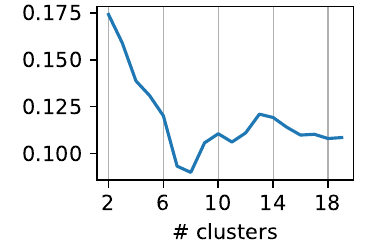}
\end{subfigure}
\begin{subfigure}[t]{0.475\linewidth}
\caption{$L_B$} 
\label{fig:sil-bible}
\includegraphics[width=\linewidth,trim={5.5cm 1cm 20cm 0.675cm},clip]{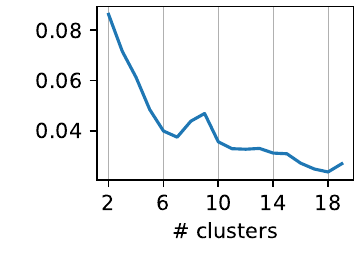}
\end{subfigure}

\caption{Silhouette analysis for the $L_T*$ embeddings trained using an initial pseudo-token (left) and the $L_B$ Bible vectors (right). Both cases present a downtrend curve with scores below 0.2. The hierarchies of $L_T*$ and $L_B$ are shown in Figures \ref{fig:ddg-ted53-pseudotoken} and \ref{fig:ddg-bible} (in the Appendix), respectively.} 
\label{fig:pseudo-token-sil}
\end{center}
\end{figure}

Figure \ref{fig:pseudo-token-sil} shows the silhouette analysis for the aforementioned embeddings ($L_T*$) together with the Bible embeddings ($L_B$) that were trained with the same configuration. We observe that the silhouette score never exceeds 0.2, and the curve keeps degrading when we examine a higher number of clusters, which contrast the trend shown in Figure \ref{fig:dendrograms}. The pattern proves that the vectors are not suitable for clustering (the hierarchies are shown in Figure \ref{fig:dendrograms-full} in the Appendix), and they might only encode enough information to perform a classification task in the multilingual NMT training and inference. For that reason, we consider it essential to use language embeddings from factors for extracting language relationships.

\section{Related work}

For language-level representations, URIEL and $\operatorname{lang2vec}$ \cite{littell-etal-2017-uriel} allow a straightforward extraction of typological binary features from different KBs. 
\citet{murawaki-2015-continuous, murawaki-2017-diachrony, murawaki-2018-analyzing} exploits them to build latent language representations with independent binary variables. 
Language features are encoded from data-driven tasks as well, such as 
NMT \cite{malaviya-etal-2017-learning} or language modelling \cite{tsvetkov-etal-2016-polyglot,ostling-tiedemann-2017-continuous,bjerva-augenstein-2018-tracking} with complementary linguistic-related target tasks \cite{bjerva-augenstein-2018-phonology}. 

Our approach is most similar to \citet{bjerva-etal-2019-probabilistic}, as they build a generative model from typological features and use language embeddings, extracted from factored language modelling at character-level, as a prior of the model to extend the language coverage. However, our method primarily differs as it is mainly based in linear algebra, encodes information from both sources since the beginning, and can deal with a small number of shared entries (e.g. 23 from $L_W$) to compute robust representations.

There has been very little work on adopting typology knowledge for NMT. There is not a deep integration of the topics \cite{ponti-etal-2019-modeling}, but one shallow and prominent case is the ranking method  \cite{lin-etal-2019-choosing} that we analysed in \S\ref{sec:broadening-clustering}. 

Finally, CCA and its variants have been previously used to derive embeddings at word-level \cite{faruqui-dyer-2014-improving,dhillon2015eigenwords,osborne-etal-2016-encoding}. \citet{kudugunta-etal-2019-investigating} also used SVCCA but to inspect sentence-level representations, where they uncover relevant insights about language similarity that are aligned with our results in \S\ref{sec:tree-inference}. However, as far as we know, this is the first time a CCA-based method has been used to compute language-level representations. 

\section{Takeaways and practical tool}

We summarise our key findings as follows:
\begin{itemize}
    \item SVCCA can fuse linguistic typology KB entries with NMT-learned embeddings without diminishing the originally encoded typological and genetic similarity of languages. 
    \item Our method is a robust alternative for identifying clusters and choosing related languages for multilingual transfer in NMT. The advantage is notable when it is not feasible to pre-train a ranking model or learn embeddings from a massive multilingual system. Assessing new languages is an important ability, given that most of them do not have even enough monolingual corpora to learn embeddings from multilingual language modelling \cite{joshi-etal-2020-state}.
    \item Factored language embeddings encode more information to agglomerate related languages than the initial pseudo-token setting. 
\end{itemize}

Furthermore, we make our code available as an open-source tool\footnote{\url{https://github.com/aoncevay/multiview-langrep}}, together with our $L_T$ factored-embeddings, to compute multi-view language representations using SVCCA. We enable the option to use other language vectors from $\operatorname{lang2vec}$ (Phonology or Phonetic Inventory) as the KB-source, and to upload new task-learned embeddings from different settings, such as one-to-many or many-to-many NMT, and also multilingual language modelling. 
Besides, given a list of languages to assess, our method will project new language representations when they are only available in the KB-view. Finally, we include the tasks of language clustering and ranking candidates, which could benefit multilingual NLP studies that involve massive datasets of hundreds of languages. 

\section{Conclusion}
We compute multi-view language representations with SVCCA using two sources: KB and NMT-learned vectors. With a typological feature prediction task and the inference of phylogenetic trees, we showed that the knowledge and language relationship encoded in both sources is preserved in the combined representation. Moreover, our approach offers important advantages because we can evaluate projected languages with entries in only one of the views and can easily extend the language coverage. The benefits are noticeable in multilingual NMT tasks, like language clustering and ranking related languages for multilingual transfer. We plan to study how to deeply incorporate our typologically-enriched embeddings in multilingual NMT, where there are promising avenues in parameter selection \cite{sachan-neubig-2018-parameter} and generation \cite{platanios-etal-2018-contextual}.

\section*{Acknowledgments}
\lettrine[image=true, lines=2, findent=1ex, nindent=0ex, loversize=.15]{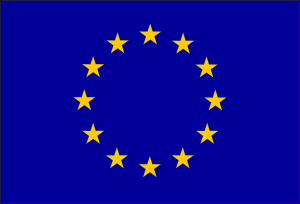}%
{T}his work was supported by funding from the European Union's Horizon 2020 research and innovation programme under grant agreements No 825299 (GoURMET) and the EPSRC fellowship grant EP/S001271/1 (MTStretch). Also, it was performed using resources provided by the Cambridge Service for Data Driven Discovery (CSD3) operated by the University of Cambridge Research Computing Service (\url{http://www.csd3.cam.ac.uk/}), provided by Dell EMC and Intel using Tier-2 funding from the Engineering and Physical Sciences Research Council (capital grant EP/P020259/1), and DiRAC funding from the Science and Technology Facilities Council (\url{www.dirac.ac.uk}). We express our thanks to Kenneth Heafield and Rico Sennrich, who provided us with access to the computing resources. 

Last but not least, we thank the organisers and participants of the First Workshop of Typology for Polyglot NLP, and the members of the Statistical Machine Translation group at the University of Edinburgh, whose provided relevant feedback in an early stage of the study.

\bibliography{emnlp2020}
\bibliographystyle{acl_natbib}

\appendix

\section{Languages and individual BLEU scores}
\label{app:languages}
We work with 53 languages pre-processed by \cite{qi-etal-2018-pre}, from where we mapped the ISO 639-1 codes to the ISO 693-2 standard. However, we need to manually correct the mapping of some codes to identify the correct language vector in the URIEL \cite{littell-etal-2017-uriel} library:
\begin{itemize}[noitemsep]
    \item \emph{zh}  (\emph{zho} , Chinese macro-language) mapped to \emph{cmn} (Mandarin Chinese).
    \item \emph{fa}  (\emph{fas} , Persian inclusive code for 11 dialects) mapped to \emph{pes} (Western/Iranian Persian).
    \item \emph{ar}  (\emph{ara} , Arabic) mapped to \emph{arb} (Standard Arabic).
\end{itemize}
We disregard working with artificial languages like Esperanto (eo) or variants like Brazilian Portuguese (pt-br) and Canadian French (fr-ca). 

Table \ref{tab:all-languages} presents the list of all the languages with the following details: ISO 693-2 code, language family, size of the training set in thousands of sentences (with their respective training size bin) and the individual BLEU score obtained per clustering approach and other baselines.

\begin{table*}[t!]
\centering
\small
\setlength\tabcolsep{3pt}
\resizebox{\linewidth}{!}{

\begin{tabular}{cll|cc|cccccccc} 
 \multicolumn{5}{c}{} & \multicolumn{8}{|c}{BLEU score per approach} \\ \hline
ISO & Language & Lang. family & Size (k) & Bin & Individual & Massive & Family &  $U_S$ & $L_T$ & $\oplus$ & SVCCA-53 & SVCCA-23 \\ \hline
 
 kaz & Kazakh & Turkic & 3 & 1 & 2.5 & 5.3 & 4.0 & 4.3 & 3.3 & 2.7 & 3.3 & 3.0 \\  
 bel & Belarusian & IE/Balto-Slavic & 4 & 1 & 3.1 & 13.0 & 14.3 & 13.7 & 4.3 & 2.8 & 12.4 & 10.1 \\  
 ben & Bengali & IE/Indo-Iranian & 4 & 1 & 3.1 & 10.5 & 5.9 & 6.2 & 4.3 & 4.6 & 4.4 & 5.7 \\  
 eus & Basque & Isolate & 5 & 1 & 2.2 & 11.1 & 2.2 & 10.9 & 5.6 & 3.9 & 6.4 & 10.1 \\  
 zlm & Malay & Austronesian & 5 & 1 & 4.1 & 20.1 & 15.6 & 19.7 & 6.5 & 4.1 & 19.6 & 19.6 \\  
 bos & Bosnian & IE/Balto-Slavic & 5 & 1 & 4.2 & 26.6 & 28.0 & 28.3 & 6.5 & 4.1 & 26.1 & 23.6 \\  
 urd & Urdu & IE/Indo-Iranian & 5 & 1 & 3.9 & 11.8 & 7.5 & 8.0 & 5.5 & 5.6 & 7.1 & 6.8 \\  
 aze & Azerbaijani & Turkic & 5 & 1 & 2.8 & 8.1 & 6.4 & 6.7 & 4.2 & 3.2 & 7.3 & 7.4 \\  
 tam & Tamil & Dravidian & 6 & 1 & 1.4 & 5.1 & 1.4 & 4.0 & 2.8 & 2.6 & 2.7 & 2.3 \\  
 mon & Mongolian & Mongolic & 7 & 1 & 2.7 & 6.9 & 2.7 & 5.7 & 3.9 & 3.5 & 5.2 & 6.1 \\  
 mar & Marathi & IE/Indo-Iranian & 9 & 1 & 3.2 & 7.0 & 5.1 & 5.2 & 4.1 & 4.0 & 3.3 & 4.7 \\  
 glg & Galician & IE/Italic & 9 & 1 & 8.4 & 24.9 & 29.1 & 26.1 & 29.0 & 28.7 & 28.9 & 28.2 \\  
 kur & Kurdish & IE/Indo-Iranian & 10 & 1 & 4.0 & 10.1 & 6.8 & 10.8 & 4.9 & 3.6 & 6.3 & 8.1 \\  
 est & Estonian & Uralic & 10 & 1 & 5.8 & 13.5 & 10.5 & 14.1 & 8.1 & 8.1 & 11.7 & 11.9 \\  
 kat & Georgian & Kartvelian & 13 & 2 & 5.8 & 14.3 & 5.8 & 14.5 & 8.8 & 4.6 & 5.6 & 5.5 \\  
 nob & Nor. Bokmal & IE/Germanic & 15 & 2 & 19.0 & 35.2 & 38.8 & 36.4 & 35.0 & 35.0 & 39.1 & 39.1 \\  
 hin & Hindi & IE/Indo-Iranian & 18 & 2 & 8.1 & 16.0 & 8.8 & 10.5 & 9.5 & 6.2 & 8.3 & 8.6 \\  
 slv & Slovenian & IE/Balto-Slavic & 19 & 2 & 8.7 & 19.5 & 19.8 & 20.2 & 21.8 & 19.3 & 18.1 & 19.7 \\  
 mya & Burmese & Sino-Tibetan & 20 & 2 & 4.9 & 10.3 & 7.6 & 7.3 & 6.0 & 4.1 & 7.7 & 3.4 \\  
 hye & Armenian & IE/Armenian & 21 & 2 & 9.0 & 16.3 & 9.0 & 16.9 & 9.8 & 13.2 & 13.3 & 12.2 \\  
 fin & Finnish & Uralic & 23 & 2 & 8.5 & 14.4 & 11.5 & 14.9 & 8.3 & 8.3 & 12.1 & 15.0 \\  
 mkd & Macedonian & IE/Balto-Slavic & 24 & 2 & 15.7 & 26.8 & 27.3 & 27.4 & 27.2 & 28.0 & 25.1 & 22.6 \\  
 lit & Lithuanian & IE/Balto-Slavic & 41 & 2 & 12.2 & 17.9 & 19.4 & 18.4 & 20.0 & 19.0 & 17.9 & 18.6 \\  
 sqi & Albanian & IE/Albanian & 43 & 2 & 20.8 & 27.8 & 20.8 & 29.1 & 28.6 & 31.6 & 26.3 & 25.8 \\  
 dan & Danish & IE/Germanic & 44 & 2 & 30.7 & 35.6 & 38.4 & 36.7 & 34.4 & 34.4 & 38.9 & 39.0 \\  
 por & Portuguese & IE/Italic & 50 & 2 & 27.2 & 32.8 & 36.9 & 33.7 & 36.6 & 36.0 & 36.7 & 36.5 \\  
 swe & Swedish & IE/Germanic & 55 & 2 & 27.0 & 30.8 & 33.6 & 31.8 & 29.7 & 29.7 & 34.3 & 34.6 \\  
 slk & Slovak & IE/Balto-Slavic & 60 & 2 & 18.1 & 24.1 & 26.0 & 24.7 & 26.8 & 25.5 & 23.7 & 22.2 \\  
 ind & Indonesian & Austronesian & 85 & 3 & 23.8 & 24.3 & 21.4 & 26.0 & 28.0 & 27.0 & 26.5 & 26.5 \\  
 tha & Thai & Kra-Dai & 96 & 3 & 15.4 & 16.8 & 15.4 & 16.9 & 19.0 & 17.6 & 17.7 & 17.7 \\  
 ces & Czech & IE/Balto-Slavic & 101 & 3 & 20.7 & 22.1 & 23.9 & 22.8 & 24.2 & 23.3 & 21.2 & 22.1 \\  
 ukr & Ukrainian & IE/Balto-Slavic & 106 & 3 & 19.8 & 20.9 & 22.6 & 22.0 & 23.5 & 22.5 & 21.2 & 21.7 \\  
 hrv & Croatian & IE/Balto-Slavic & 120 & 3 & 28.5 & 27.5 & 30.4 & 28.9 & 30.8 & 31.5 & 28.3 & 26.7 \\  
 ell & Greek & IE/Hellenic & 132 & 3 & 31.9 & 29.9 & 31.9 & 30.9 & 32.2 & 33.4 & 34.2 & 32.7 \\  
 srp & Serbian & IE/Balto-Slavic & 134 & 3 & 26.4 & 25.6 & 28.3 & 27.1 & 28.8 & 29.4 & 26.3 & 25.4 \\  
 hun & Hungarian & Uralic & 145 & 3 & 19.1 & 17.2 & 17.0 & 17.9 & 21.3 & 17.7 & 18.0 & 18.7 \\  
 fas & Persian & IE/Indo-Iranian & 148 & 3 & 20.9 & 18.5 & 9.0 & 19.7 & 22.4 & 22.2 & 8.4 & 17.9 \\  
 deu & German & IE/Germanic & 165 & 3 & 30.1 & 25.5 & 29.5 & 26.9 & 31.4 & 31.7 & 29.9 & 29.6 \\  
 vie & Vietnamese & Austroasiatic & 169 & 3 & 22.7 & 20.3 & 22.7 & 21.6 & 23.6 & 22.2 & 22.3 & 22.3 \\  
 bul & Bulgarian & IE/Balto-Slavic & 172 & 3 & 33.9 & 29.9 & 31.9 & 31.4 & 33.3 & 33.1 & 34.2 & 33.8 \\  
 pol & Polish & IE/Balto-Slavic & 173 & 3 & 18.9 & 17.4 & 19.1 & 18.2 & 19.3 & 18.9 & 18.3 & 16.9 \\  
 ron & Romanian & IE/Italic & 178 & 4 & 30.0 & 25.8 & 30.7 & 27.0 & 28.1 & 30.8 & 30.8 & 29.6 \\  
 tur & Turkish & Turkic & 179 & 4 & 19.5 & 14.6 & 16.2 & 15.6 & 20.7 & 20.3 & 17.1 & 17.9 \\  
 nld & Dutch & IE/Germanic & 181 & 4 & 31.7 & 26.6 & 30.6 & 27.7 & 32.5 & 33.0 & 31.2 & 30.5 \\  
 fra & French & IE/Italic & 189 & 4 & 35.6 & 30.6 & 35.9 & 32.0 & 35.9 & 36.1 & 34.3 & 34.5 \\  
 spa & Spanish & IE/Italic & 193 & 4 & 37.2 & 32.2 & 37.4 & 33.5 & 37.5 & 37.0 & 37.5 & 36.2 \\  
 cmn & Chinese & Sino-Tibetan & 197 & 4 & 14.9 & 13.5 & 13.9 & 12.6 & 15.8 & 15.8 & 14.7 & 14.7 \\  
 jpn & Japanese & Japonic & 201 & 4 & 9.8 & 8.5 & 9.8 & 8.6 & 10.8 & 10.8 & 9.8 & 9.7 \\  
 ita & Italian & IE/Italic & 201 & 4 & 33.6 & 28.6 & 34.1 & 29.6 & 33.9 & 33.3 & 33.7 & 32.4 \\  
 kor & Korean & Koreanic & 202 & 4 & 14.4 & 12.2 & 14.4 & 11.9 & 15.1 & 15.0 & 13.3 & 5.8 \\  
 rus & Russian & IE/Balto-Slavic & 205 & 4 & 20.4 & 18.1 & 19.4 & 19.0 & 20.1 & 19.5 & 18.3 & 18.8 \\  
 heb & Hebrew & Afroasiatic & 208 & 4 & 32.4 & 24.4 & 32.9 & 25.8 & 29.9 & 30.3 & 31.9 & 31.6 \\  
 arb & Arabic & Afroasiatic & 211 & 4 & 26.5 & 20.5 & 27.5 & 21.6 & 25.4 & 26.5 & 27.5 & 26.6 \\  
 
\hline
\multicolumn{5}{r}{Average $\rightarrow$} & 16.7 & 19.8 & 19.8 & 20.0 & 19.6 & 19.2 & 20.0 & 19.8 \\  
\hline

\end{tabular}
}
\caption{List of languages with their BLEU scores per clustering approach (IE=Indo-European).} 
\label{tab:all-languages}
\end{table*}

\section{Correlation of SVCCA with genetic similarity}
\label{app:lexical-sim} 
\citet{Bjerva2019-bl} argued that raw language embeddings from language modelling correlates with \emph{genetic} and structural similarity\footnote{We note that \citet{Bjerva2019-bl} used monolingual texts translated from different languages to investigate what kind of genetic information is preserved. Concerning structural similarity, they computed a distance matrix using syntax-dependency-tags counts per language from annotated treebanks. We leave this analysis for further work.}. For the former, they correlated a distance matrix with pairwise-leaf-distances of the GS. However, \citet{Serva2008-fb} originally inferred the phylogeny by comparing the translated Swadesh list of 200-words \cite{dyen1992indoeuropean} with Levenshtein (edit) distance. The list is a crafted set of concepts for comparative linguistics (e.g. I, eye, sleep), and it is usually processed by lexicostatistics methods to study language relationship through time. Therefore, 
we prefer to argue that corpus-based embeddings could partially encode \textbf{lexical} similarity of languages. 

We perform a Spearman correlation between the cophenetic matrix\footnote{ Pairwise-distances of the hierarchy's leaves (languages).} of the GS and the pairwise cosine-distance matrices of $U_S$, $L_T$ and SVCCA($U_S$,$L_T$), where we obtain correlation coefficients of 0.48, 0.68 and 0.80, respectively (p-values$<$0.001). Our conclusion is that typological knowledge strengthen the representation of lexical similarity within NMT-learned embeddings.

\section{SVD explained variance selection}
\label{app:parameter-search}

To compute SVCCA, we transform each source space using SVD, where we can choose to preserve a number of dimensions that represents an accumulated explained variance of the original dataset. For that reason, we perform a parameter sweep between 0.5 and 1.0 using 0.05 incremental steps. For a fair comparison, we also transform the single spaces (KB or Learned) with SVD and look for the optimal threshold. 

\paragraph{Prediction of typological features. } We selected a 0.5 threshold for the NMT-learned vectors of $L_B$ and $L_W$, and 0.7 for $L_T$. In case of the SVCCA representation, $L_T$ uses [0.75,0.70], whereas $L_B$ and $L_W$ employ [0.95,0.50] values. The parameter values are for both one-language-out and one-family-out settings. 
We can argue that there is redundancy in the NMT-learned embeddings, as the prediction of typological features with Logistic Regression always prefers a dimensionality-reduced version instead of the original data (threshold at 1.0).

\paragraph{Language phylogeny inference.} In Table \ref{tab:tree-TED-multi-views-with-ths}, we report the optimal value for the SVD explained variance ratio in each single and multi-view (concatenation and SVCCA) setting.

\paragraph{Language clustering (and ranking).}
 We cannot perform an exhaustive analysis for the threshold of the explained variance ratio per view. As our main goal is to increase the coverage of languages steadily, we must determine what configuration allows a stable growth of the hierarchy. 
 
\begin{figure}[t!]
\begin{center}

\begin{subfigure}[t]{0.45\linewidth}
\caption{Syntax: $U_S$ \textsubscript{(0.65)}} 
\label{fig:bootstrap-syntax}
\includegraphics[width=\linewidth,trim={38.5cm 0.3cm 88.5cm 0.735cm},clip]{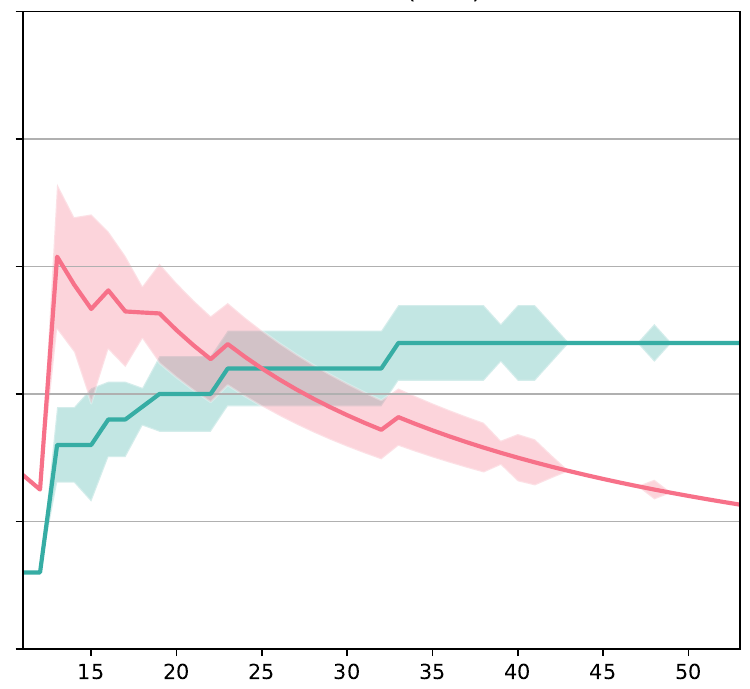} 
\end{subfigure}
\begin{subfigure}[t]{0.45\linewidth}
\caption{TED-53: $L_T$ \textsubscript{(0.6)}} 
\label{fig:bootstrap-learned}
\includegraphics[width=\linewidth,trim={26cm 0.3cm 101cm 0.735cm},clip]{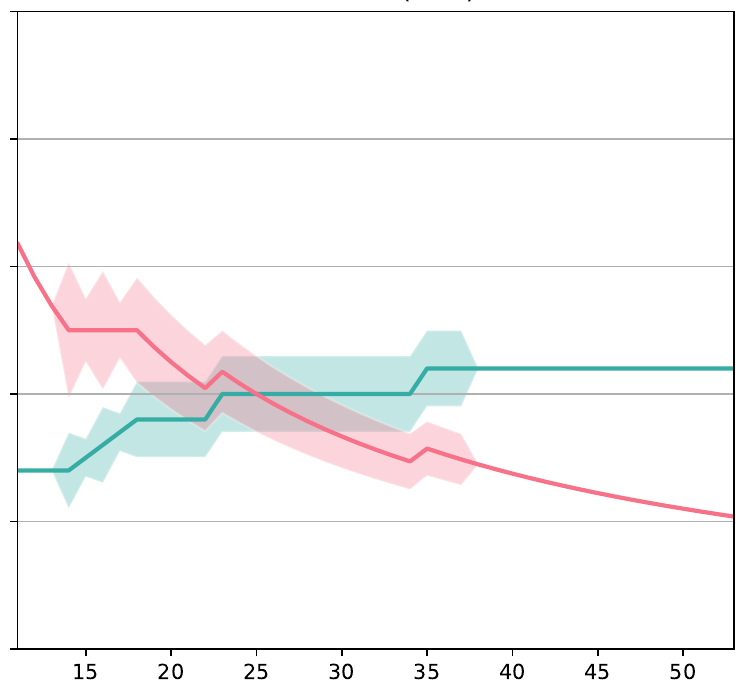}
\end{subfigure}

\caption{Analysis of the number of clusters (blue) and the ratio of number of clusters per total languages (red) given the chosen thresholds of explained variance ratio. We show the confidence interval computed from the bootstrapping, and we observe that the number of clusters is stable since 42 and 38 languages for $U_S$ and $L_T$ vectors, respectively. }
\label{fig:variance-search}
\end{center}
\end{figure}

 We thereupon take inspiration from bootstrap clustering \cite{nerbonne2008projecting},
 and increase the number of language entries from few entries (e.g. 10) to 53 by resample bootstrapping using each of the source vectors: $U_S$, $L_T$ and $L_W$. Afterwards, we search for the threshold value that preserves a stable number of clusters given the peak silhouette value. Our heuristic looks for the least variability throughout the incremental bootstrapping (Fig. \ref{fig:variance-search}). 
 
 We found that 0.65 is the most stable value for $U_S$, whereas 0.60 is the best one for both $L_T$ and $L_W$, so we thereupon fix SVCCA-53 and SVCCA-23 to [0.65,0.6]. We also apply the chosen thresholds on the concatenation baseline for a fair comparison. In the single-view cases, the transformations with the tuned variance ratio do not overcome any non-optimised counterparts.

\begin{table}[t!]
\centering
\setlength\tabcolsep{4pt}
\resizebox{\linewidth}{!}{%
\begin{tabular}{l|c|cc}
                     & Single & $U_S\oplus L_*$  & SVCCA($U_S$,$L_*$)  \\ \hline
$U_S$ (Syntax)  & 30 / 0.45 \textsubscript{(0.5)} \\                    
$L_B$ (Bible)   & 35 / 0.54 \textsubscript{(0.9)} & 27 / 0.42 \textsubscript{(0.70,0.55)} & 23 / 0.34 \textsubscript{(0.70,0.75)} \\
$L_W$ (WIT-23)  & 35 / 0.62 \textsubscript{(0.8)} & 23 / 0.41 \textsubscript{(0.75,0.95)} & 27 / 0.48 \textsubscript{(0.50,0.95)} \\
$L_T$ (TED-53)  & 15 / 0.26 \textsubscript{(0.6)} & 18 / 0.29 \textsubscript{(0.70,0.55)} & \textbf{10} / \textbf{0.15} \textsubscript{(1.00,0.55)} \\ \hline
\end{tabular}
}
\caption{Similar to Table \ref{tab:tree-TED-multi-views}, but including the optimal values for the SVD explained variance in each setting.} 
\label{tab:tree-TED-multi-views-with-ths}
\end{table}

\section{Language clustering results by language families}
\label{app:families}

\begin{table*}[t!]
\centering
\small
\setlength\tabcolsep{4pt}
\resizebox{.9\linewidth}{!}{

\begin{tabular}{l|c@{\hskip 3pt}c|c@{\hskip 3.5pt}c@{\hskip 4pt}cccc |r@{\hskip 2pt}l|r@{\hskip 2pt}l}
Lang. families & \# L & Size (k) & Individual & Massive & Family &  $U_S$ & $L_T$ & $\oplus$ & \multicolumn{2}{c|}{\textbf{SVCCA-53}} & \multicolumn{2}{c}{\textbf{SVCCA-23}} \\ \hline

Isolate (Basque)	&	1	&	5	&	2.20	&	\textbf{11.10}	&	2.20	&	\emph{10.90}	&	5.60	&	3.90	&	6.40 & \textsubscript{\textDelta-4.7}	&	10.10 & \textsubscript{\textDelta-1.0}  \\
Dravidian	&	1	&	6	&	1.40	&	\textbf{5.10}	&	1.40	&	\emph{4.00}	&	2.80	&	2.60	&	2.70 & \textsubscript{\textDelta-2.4}	&	2.30 & \textsubscript{\textDelta-2.8}  \\
Mongolic	&	1	&	7	&	2.70	&	\textbf{6.90}	&	2.70	&	5.70	&	3.90	&	3.50	&	5.20 & \textsubscript{\textDelta-1.7}	&	\emph{6.10} & \textsubscript{\textDelta-0.8}  \\
Kartvelian	&	1	&	13	&	5.80	&	\emph{14.30}	&	5.80	&	\textbf{14.50}	&	8.80	&	4.60	&	5.60 & \textsubscript{\textDelta-8.9}	&	5.50 & \textsubscript{\textDelta-9.0}  \\
IE/Armenian	&	1	&	21	&	9.00	&	\emph{16.30}	&	9.00	&	\textbf{16.90}	&	9.80	&	13.20	&	13.30 & \textsubscript{\textDelta-3.6}	&	12.20 & \textsubscript{\textDelta-4.7}  \\
IE/Albanian	&	1	&	44	&	20.80	&	27.80	&	20.80	&	\emph{29.10}	&	28.60	&	\textbf{31.60}	&	26.30 & \textsubscript{\textDelta-5.3}	&	25.80 & \textsubscript{\textDelta-5.8}  \\
Kra-Dai	&	1	&	97	&	15.40	&	16.80	&	15.40	&	16.90	&	\textbf{19.00}	&	17.60	&	\emph{17.70} & \textsubscript{\textDelta-1.3}	&	\emph{17.70} & \textsubscript{\textDelta-1.3}  \\
IE/Hellenic	&	1	&	132	&	31.90	&	29.90	&	31.90	&	30.90	&	32.20	&	\emph{33.40}	&	\textbf{34.20} &		&	32.70 & \textsubscript{\textDelta-1.5}  \\
Austroasiatic	&	1	&	170	&	\emph{22.70}	&	20.30	&	\emph{22.70}	&	21.60	&	\textbf{23.60}	&	22.20	&	22.30 & \textsubscript{\textDelta-1.3}	&	22.30 & \textsubscript{\textDelta-1.3}  \\
Japonic	&	1	&	201	&	9.80	&	8.50	&	9.80	&	8.60	&	\textbf{10.80}	&	\textbf{10.80}	&	9.80 & \textsubscript{\textDelta-1.0}	&	9.70 & \textsubscript{\textDelta-1.1}  \\
Koreanic	&	1	&	203	&	14.40	&	12.20	&	14.40	&	11.90	&	\textbf{15.10}	&	\emph{15.00}	&	13.30 & \textsubscript{\textDelta-1.8}	&	5.80 & \textsubscript{\textDelta-9.3}  \\
Austronesian	&	2	&	91	&	13.95	&	22.20	&	18.50	&	22.85	&	17.25	&	15.55	&	\textbf{23.05} &		&	\textbf{23.05} &	  \\
Sino-Tibetan	&	2	&	218	&	9.90	&	\textbf{11.90}	&	10.75	&	9.95	&	10.90	&	9.95	&	\emph{11.20} & \textsubscript{\textDelta-0.7}	&	9.05 & \textsubscript{\textDelta-2.8}  \\
Afroasiatic	&	2	&	420	&	29.45	&	22.45	&	\textbf{30.20}	&	23.70	&	27.65	&	28.40	&	\emph{29.70} & \textsubscript{\textDelta-0.5}	&	29.10 & \textsubscript{\textDelta-1.1}  \\
Uralic	&	3	&	180	&	11.13	&	15.03	&	13.00	&	\textbf{15.63}	&	12.57	&	11.37	&	13.93 & \textsubscript{\textDelta-1.7}	&	\emph{15.20} & \textsubscript{\textDelta-0.4}  \\
Turkic	&	3	&	189	&	8.27	&	9.33	&	8.87	&	8.87	&	\emph{9.40}	&	8.73	&	9.23 & \textsubscript{\textDelta-0.2}	&	\textbf{9.43} &	  \\
IE/Germanic	&	5	&	462	&	27.70	&	30.74	&	34.18	&	31.90	&	32.60	&	32.76	&	\textbf{34.68} &		&	\emph{34.56} & \textsubscript{\textDelta-0.1}  \\
IE/Indo-Iranian	&	6	&	198	&	7.20	&	\textbf{12.32}	&	7.18	&	\emph{10.07}	&	8.45	&	7.70	&	6.30 & \textsubscript{\textDelta-6.0}	&	8.63 & \textsubscript{\textDelta-3.7}  \\
IE/Italic	&	6	&	823	&	28.67	&	29.15	&	\textbf{34.02}	&	30.32	&	33.50	&	\emph{33.65}	&	\emph{33.65} & \textsubscript{\textDelta-0.4}	&	32.90 & \textsubscript{\textDelta-1.1}  \\
IE/Balto-Slavic	&	13	&	1,171	&	17.74	&	22.26	&	\textbf{23.88}	&	\emph{23.24}	&	22.05	&	21.30	&	22.39 & \textsubscript{\textDelta-1.5}	&	21.71 & \textsubscript{\textDelta-2.2}  \\

\hline
\multicolumn{3}{r|}{Weighted average $\rightarrow$}   &	16.70	&	19.76	&	19.79	&	\textbf{20.03}	&	19.60	&	19.16	&	\emph{19.97} & \textsubscript{\textDelta-0.1}	&	19.82 & \textsubscript{\textDelta-0.1}  \\
\multicolumn{3}{r|}{Number of clusters/models $\rightarrow$}	&   53	&   1	&   20	&   3	&   11	&   18	&   10	&  & 10 \\
\hline
\end{tabular}
}
\caption{BLEU score average per language family (IE=Indo-European). Every method includes the weighted BLEU average per number of languages (\#L) and the number of clusters/models. Bold and italic represent first and second best results per family. \textDelta~for SVCCA indicates the difference with respect to the highest score.} 
\label{tab:fam-ted53}
\end{table*}

Following a guide for evaluating multilingual benchmarks \cite{anastasopoulos-2019-multilingual}, we also group the scores by language families. Table \ref{tab:fam-ted53} includes the overall weighted average per number of languages in each family branch. 
We observe that most of the approaches have obtained clusters with similar overall translation accuracy. The individual models are the only ones that significantly underperform. 
The poor performance is transferred to the Family baseline, as most of the groups contains only one language given the low language diversity of the dataset.

The $U_S$ vectors obtain the highest overall accuracy, mostly from their few large clusters (see Fig. \ref{fig:ddg-syntax}). Meanwhile, SVCCA-53 achieves the second-best overall result, by a minimal margin, and with 3 to 7 languages per cluster, which are usually faster to converge. 
Besides, the massive model, the $L_T$ embeddings and the concatenation baseline present a competitive achievement as well. However, the first requires more resources to train until convergence, whereas the last two need the 53 pre-trained embeddings from a previous massive system. 

In contrast, SVCCA-23 is a faster alternative if we want to target specific new languages (see Fig. \ref{fig:ddg-svcca23}). We only require a small group of language embeddings (e.g. $L_W$ of 23 entries) and project the rest with SVCCA and a set KB-vectors as a side view. For instance, if we need to deploy a translation model for Basque or Thai, we could reach a comparable or better accuracy to a massive model with the SVCCA-23 chosen clusters of only 3 (Arabic, Hebrew) or 5 (Chinese, Indonesian, Vietnamese, Malay) languages, respectively. 

\begin{figure*}[t!]
\begin{center}
\centering

\begin{subfigure}[t]{\linewidth}
\caption{$L_B$: NMT-learned from Bible} 
\label{fig:ddg-bible}
\includegraphics[width=\linewidth,clip]{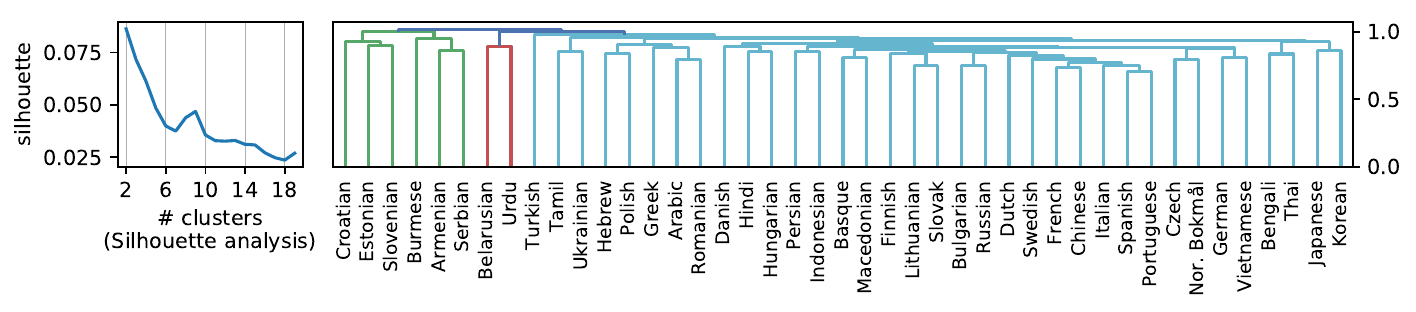}
\end{subfigure}

\begin{subfigure}[t]{\linewidth}
\caption{$L_T*$: NMT-learned from TED-53 but with initial pseudo-tokens} 
\label{fig:ddg-ted53-pseudotoken}
\includegraphics[width=\linewidth,clip]{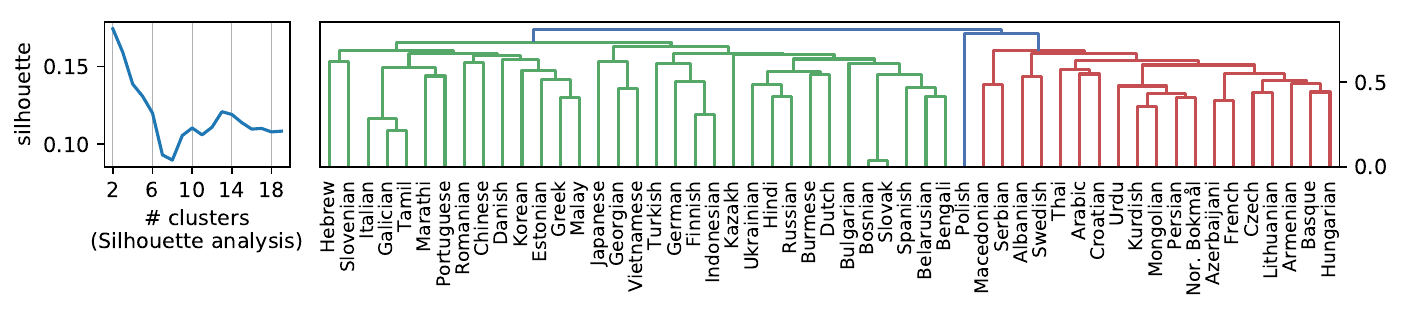}
\end{subfigure}

\begin{subfigure}[t]{\linewidth}
\caption{SVCCA-53*($U_S$,$L_T*$): SVCCA representations of Syntax and TED-53 but with initial pseudo-tokens} 
\label{fig:ddg-svcca53-pseudotoken}
\includegraphics[width=\linewidth,clip]{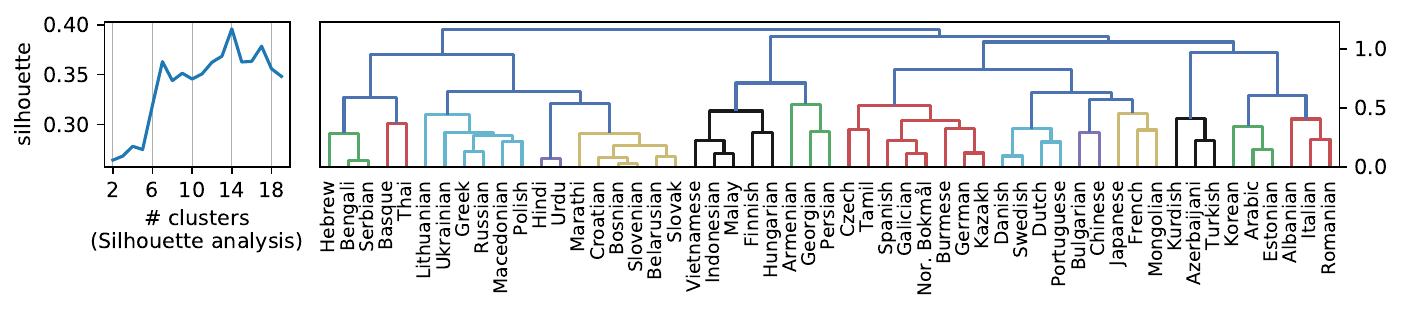}
\end{subfigure}

\caption{Silhouette analysis and dendrograms for clustering the 53 languages of TED-53 using different language representations. In (a) and (b), we note that the silhouette score is below 0.2 (1 is best), and the hierarchies do not define natural groups for the languages, as they are usually very separated from each other. In (c), we note that SVCCA is affected by the noisy agglomeration of the original NMT-learned embeddings with initial pseudo-tokens.} 
\label{fig:dendrograms-full}
\end{center}
\end{figure*}

\end{document}